\documentclass{article}

\usepackage{genai4health_2026}
\usepackage[utf8]{inputenc}
\usepackage[T1]{fontenc}
\usepackage{url}
\usepackage{booktabs}
\usepackage{amsmath,amssymb,amsfonts}
\usepackage{graphicx}
\usepackage{xcolor}
\usepackage{multirow}
\usepackage{array}
\usepackage{float}
\usepackage{tikz}
\usetikzlibrary{arrows.meta,positioning}
\usepackage[hidelinks]{hyperref}

\newcommand{\method}{\textsc{RSP}}
\newcommand{\dataset}[1]{\textsc{#1}}
\newcommand{\randomfour}{Random@\(4B\)}

\title{An Empirical Study of Development-Tuned Representation Sensitivity for Medical QA Data Selection}

\author{
  Yuexin Wu \quad Dayou Yu \quad Vasile Rus \\
  Department of Computer Science, University of Memphis \\
  \texttt{ywu10@memphis.edu \quad dayou.yu@memphis.edu \quad vrus@memphis.edu}
}

\begin{document}
\maketitle

\begin{abstract}
We study medical question-answering (QA) data selection when a developer already has questions, labels, and auxiliary explanatory texts but can retain only \(B\) examples for a final LoRA update. Representation-Sensitivity Prioritization (\method{}) first restricts the candidate bank to its top-\(4B\) predictive-entropy examples, then combines the frequency of answer flips with non-flipping contraction of the top-two answer margin under hidden-state perturbations. We evaluate this operational score on five medical QA datasets with \texttt{google/medgemma-4b-it}, \(B=128\), and three seed branches. After removing four MedExpQA test questions that also occur in its acquisition pool, dataset-specific development-tuned \method{} attains the highest observed five-dataset mean accuracy (53.44\%) and Macro-F1 (48.21\%) among the evaluated selectors. Relative to random selection within the same entropy-restricted pool, the descriptive differences are +1.49 and +2.10 percentage points, respectively; paired 95\% bootstrap intervals include zero. The audit mixes benchmarks with different relationships to documented backbone post-training sources, and the same test cohorts had been used in earlier analyses before the expanded tuning grid was designed. Results are therefore exploratory, modest, and dataset-dependent rather than evidence of statistical superiority. \method{} point estimates also exceed an archived no-update reference, but unpinned model revisions and incomplete historical input identifiers limit attribution of that difference to fine-tuning. Because every selected training example includes auxiliary text and no matched answer-only control is evaluated, the experiments also do not identify a causal benefit from auxiliary-text conditioning itself.
\end{abstract}

\section{Introduction}
\label{sec:introduction}

Developers adapting a medical foundation model may already possess an authorized bank of medical questions, answer labels, and auxiliary explanatory texts, while a planned update can retain only a fixed number of training examples. This creates an offline curation problem: which \emph{auxiliary-text-bearing examples} should enter the final update? Here, \(B\) represents an externally set limit on examples approved for a scheduled update, for example a cap on case-by-case expert re-review before an update is deployed. Such a cap can coexist with a larger labeled bank. This is a motivating scenario, not an evaluated review workflow: the retained-set budget does not measure expert effort or equalize the cost of creating, validating, scoring, or tuning the bank, and auxiliary fields can vary substantially in source and quality.

Common selection heuristics emphasize output uncertainty or representation coverage. Entropy and margin identify uncertain predictions; BALD-style criteria use stochastic disagreement \citep{houlsby2011bayesian,kirsch2019batchbald}; and diversity methods seek broader feature-space coverage \citep{hubotter2025sift}. These signals do not directly ask how a local change to a question representation affects the answer boundary. Representation-Sensitivity Prioritization (\method{}) instead records whether perturbing the final decoder block changes the predicted answer and, when it does not, how much the top-two answer margin contracts. The motivating hypothesis is that local answer-boundary fragility may help identify examples useful for adaptation. High sensitivity may also occur on ambiguous questions, and label-blind scoring cannot exclude mislabeled examples. The response depends on the perturbation scale and is not a validated measure of training value or clinical data quality. We evaluate an operational heuristic; the experiments do not identify a mechanism linking sensitivity to training utility.

Our goal is to examine whether this operational sensitivity signal is useful under a fixed protocol, not to establish a universally optimal selector. Candidate-level \method{} scoring is auxiliary-text-blind and label-blind, but its scale coefficient is chosen using labeled development performance. The complete selector is therefore development-tuned rather than label-free. We make three contributions:

\begin{itemize}
    \item We define a common medical-QA curation protocol in which \(B=128\) is the final retained-set budget, not an annotation-cost or total-compute budget.
    \item We specify and evaluate the flip--contraction \method{} score, including its exact representation, answer-probability, tie-breaking, and tuning procedures.
    \item Across five benchmarks, we report descriptive point-estimate trends, paired uncertainty from retained predictions, dataset-level heterogeneity, development-search cost, and the boundaries of the resulting claim.
\end{itemize}

\paragraph{Intended use.}
We study offline post-training data curation for a medical foundation model. The intended setting is a developer who already has an authorized bank of medical questions, labels, and auxiliary explanatory texts but can include only a small subset in a fixed LoRA update. The experiments evaluate medical-knowledge QA benchmarks rather than patient-level clinical decision making, and the results do not establish clinical safety or deployment readiness.

\section{Related Work}
\label{sec:related}

\paragraph{Explanatory supervision.}
Chain-of-thought (CoT) prompting can expose useful intermediate reasoning on multi-step tasks \citep{wei2022chain}, but a generated or collected explanation need not faithfully reflect the process that produced an answer \citep{turpin2023unfaithful,lanham2023faithfulness}. Our datasets contain heterogeneous auxiliary text---including reference explanations and an abstract conclusion---so we avoid treating every field as human CoT. We select examples from a pre-existing bank and supervise only the final answer.

\paragraph{Data selection.}
Entropy, margin, BALD, and BatchBALD quantify different forms of uncertainty \citep{houlsby2011bayesian,kirsch2019batchbald}. SIFT combines uncertainty preselection with representation-space diversity \citep{hubotter2025sift}. Recent reasoning-data selectors include rank--surprisal alignment (RSR), step-length-corrected likelihood (ASLEC-DROP), neuron-activation signals (NAIT), and loss behavior around perturbed checkpoints (TEMP) \citep{yang2026rsr,wang2026aslec,chen2026nait,jin2026temp}. Because our implementations adapt these methods to a fixed, already-labeled bank and a shared LoRA endpoint, we identify the adapted variants explicitly rather than presenting them as exact reproductions.

\paragraph{Perturbation signals.}
Prediction stability under noise complements uncertainty and diversity. Our Noise-Stability comparator constructs a hidden-sensitivity representation before diversity selection. \method{} instead preserves the direction of the answer-boundary response by separating discrete flips from non-flipping margin contraction. We treat both as operational heuristics, not calibrated measures of uncertainty or perturbation severity.

\section{Development-Tuned Representation Sensitivity}
\label{sec:method}

\subsection{Problem Setup}

Let \(\mathcal{C}=\{(x_i,y_i,r_i)\}_{i=1}^{N}\) be a bank containing a question and its options \(x_i\), an answer-position label \(y_i\), and auxiliary explanatory text \(r_i\). A selector returns \(\mathcal{S}\subset\mathcal{C}\), \(|\mathcal{S}|=B\). Auxiliary-text-blind candidate scoring may use the question, options, and domain, whereas auxiliary-text-aware scoring may also inspect \(r_i\). The UID is used only for bookkeeping and selection-signature checks; it is not a semantic model input. Every selected training example contains \(r_i\) in its context and places loss only on final-answer target tokens. Evaluation presents \(x_i\) without \(r_i\).

The candidate-level \method{} calculation never reads \(y_i\) or \(r_i\). However, labeled development accuracy and Macro-F1 are later used to choose \(\lambda\), so the full procedure uses development labels. The item budget describes only the final retained set and does not price explanation creation, score computation, hyperparameter search, or example length.

\subsection{Representation-Sensitivity Prioritization}

Scoring starts from the unadapted \texttt{google/medgemma-4b-it} checkpoint; no warm-up adapter is loaded. For item \(i\), let \(H_i^{(L)}\in\mathbb{R}^{T_i\times d}\) be the complete output tensor of the final decoder block, before subsequent model-level normalization and the language-model head. Each stochastic pass adds independent absolute Gaussian noise to every sequence position and hidden dimension:
\begin{equation}
\widetilde{H}_i^{(L,k)}=H_i^{(L)}+\epsilon_i^{(k)},
\qquad
\epsilon_i^{(k)}\sim\mathcal{N}(0,\sigma^2 I).
\label{eq:perturb}
\end{equation}
Although noise is instantiated over the stored final-block output tensor, subsequent normalization and the language-model head are token-wise, and the score uses next-token logits at the final prompt position. Consequently, only perturbation at that position affects the restricted answer probabilities; nonfinal-position noise cannot propagate through later attention because the intervention follows the final decoder block. Operationally, this is a final-position output-representation perturbation. Let \(z_{i,T_i}\) be the final-position vocabulary logits, \(\mathcal{A}_i\) the legal option letters, and \(v_y\) the single tokenizer ID for the string \texttt{" y"}. The clean distribution is
\begin{equation}
p_i(y)=\frac{\exp z_{i,T_i}(v_y)}
{\sum_{c\in\mathcal{A}_i}\exp z_{i,T_i}(v_c)},
\qquad y\in\mathcal{A}_i,
\label{eq:option-prob}
\end{equation}
and \(\widetilde p_i^{(k)}\) is defined analogously after Eq.~\eqref{eq:perturb}. This is a restricted next-token softmax over legal option letters, not a sequence probability for the option text. The implementation requires each leading-space letter to be one token; multi-token answer labels are unsupported.

Let \(\widehat y_i=\arg\max_y p_i(y)\). Exact probability ties resolve to the earliest legal letter (A before B, and so on). We define the flip rate
\begin{equation}
\operatorname{Flip}(x_i)=\frac{1}{K}\sum_{k=1}^{K}
\mathbf{1}\!\left[\arg\max_y\widetilde p_i^{(k)}(y)\ne\widehat y_i\right].
\label{eq:flip}
\end{equation}
If \(p_{i,(1)}\) and \(p_{i,(2)}\) are the two largest clean probabilities, then \(m_i^{(0)}=p_{i,(1)}-p_{i,(2)}\). For pass \(k\), the clean winner's margin is
\begin{equation}
m_i^{(k)}=\widetilde p_i^{(k)}(\widehat y_i)
-\max_{y\ne\widehat y_i}\widetilde p_i^{(k)}(y),
\end{equation}
and the non-flipping contraction is
\begin{equation}
\operatorname{Drop}(x_i)=\frac{1}{K}\sum_{k=1}^{K}
\mathbf{1}\!\left[\arg\max_y\widetilde p_i^{(k)}(y)=\widehat y_i\right]
\max\!\left(0,m_i^{(0)}-m_i^{(k)}\right).
\label{eq:drop}
\end{equation}
The priority score is
\begin{equation}
s_{\lambda}(x_i)=\operatorname{Flip}(x_i)
+\lambda\operatorname{Drop}(x_i).
\label{eq:rsp}
\end{equation}
A larger score has higher priority. We first retain the \(\min(4B,N)\) examples with highest predictive entropy, using a stable sort, then rank that pool by Eq.~\eqref{eq:rsp} and keep the top \(B\). Exact score ties resolve by original pool-row index. With \(K=8\), Flip has nine values; Drop refines ties when \(\lambda>0\), while the row-index rule can be consequential when \(\lambda=0\). For each dataset--seed pair, all \(\lambda\) values reuse the same eight perturbation draws. The score is an operational, development-tuned flip--contraction mixture rather than a calibrated or monotone measure of perturbation severity.

\subsection{Dataset-Specific Development Tuning}

We fix \(K=8\), the final decoder block, and \(\sigma=0.02\). The expanded development grid spans 0 to 1024 (Appendix~\ref{app:lambda-grid}); settings producing an already-seen 128-item selection signature are not retrained. For each dataset, we choose \(\lambda\) by mean development accuracy over seeds 13, 23, and 37, with development Macro-F1 and then smaller \(\lambda\) as tie breakers. Within the expanded run, only development metrics choose the winner and test predictions are then obtained for that frozen winner. However, the same test cohorts had been evaluated in earlier analyses before the expanded search was designed, so these results are retrospective and descriptive rather than confirmatory held-out estimates. The selected coefficients are 0.25 (AfriMed-QA), 448 (MedExpQA), 24 (MedExQA), 0 (PubMedQA), and 56 (MedMCQA). Their range is evidence of tuning dependence, not a transferable numeric scale. For a new dataset, these results provide no validated default coefficient: applying the studied procedure requires a labeled development set and a dataset-specific search, and does not guarantee an improvement.

\begin{figure*}[t]
\centering
\begin{tikzpicture}[
    font=\scriptsize,
    box/.style={draw, rounded corners=2pt, align=center, text width=2.65cm, minimum height=1.05cm, inner sep=3pt},
    blind/.style={box, fill=blue!7},
    train/.style={box, fill=orange!10},
    arrow/.style={-{Latex[length=2mm]}, thick}
]
\node[box] (bank) {Candidate bank\\\(x_i,y_i,r_i\)};
\node[blind, right=4mm of bank] (entropy) {Raw backbone\\entropy top-\(\min(4B,N)\)};
\node[blind, right=4mm of entropy] (noise) {\(K=8\) perturbed\\answer distributions};
\node[blind, right=4mm of noise] (score) {Flip/Drop score\\top-\(B\) examples};
\node[train, below=7mm of score] (lora) {LoRA training\\\([x_i,r_i]\rightarrow y_i\)};
\node[train, left=4mm of lora] (loss) {Auxiliary text in context\\loss only on answer tokens};
\node[box, left=4mm of loss] (eval) {Evaluation\\\(x_i\rightarrow y_i\), no auxiliary text};
\draw[arrow] (bank) -- node[above, font=\tiny]{sanitize} (entropy);
\draw[arrow] (entropy) -- (noise);
\draw[arrow] (noise) -- (score);
\draw[arrow] (score) -- (lora);
\draw[arrow] (lora) -- (loss);
\draw[arrow] (loss) -- (eval);
\node[font=\scriptsize, text=blue!55!black, align=center, text width=7cm, above=1.5mm of noise] {Candidate scoring reads question/options, not \(y_i\) or \(r_i\)};
\end{tikzpicture}
\caption{Study protocol for \method{}. Candidate-level \method{} scoring is label- and auxiliary-text-blind; auxiliary text is introduced after \method{} selection in the LoRA training context. Auxiliary-text-aware comparator protocols differ as documented in Appendix~\ref{app:baseline-table}.}
\label{fig:pipeline}
\end{figure*}

\section{Experimental Setup}
\label{sec:experiments}

\subsection{Datasets and Auxiliary-Text Provenance}

Table~\ref{tab:datasets} gives the fixed internal partitions and the exact auxiliary field used. All normalized records have nonempty questions, options, labels, and auxiliary text, and the internal UID sets are disjoint. Internal disjointness does not imply independence from backbone post-training or adherence to an original benchmark split.

\begin{table*}[t]
\centering
\caption{Datasets and fixed internal partitions. ``MG train family'' means the \dataset{MedGemma} report documents using that benchmark's training split; exact item overlap cannot be computed because its item manifest is unavailable.}
\label{tab:datasets}
\scriptsize
\setlength{\tabcolsep}{1.8pt}
\begin{tabular}{@{}lp{1.45cm}p{2.85cm}p{3.15cm}p{2.35cm}rrr@{}}
\toprule
Dataset & Answer format & Auxiliary text used; provenance & Pool/dev/test construction & Backbone-overlap status & Pool & Dev & Test \\
\midrule
AfriMed-QA & 5-option MCQ & \texttt{answer\_rationale}; dataset-provided, crowdsourced human text & Official train shuffled into pool/dev; official test retained & MG train family documented; study test from official test & 1,500 & 559 & 892 \\
MedExpQA & 4/5-option MCQ & \texttt{full\_answer\_no\_ref}; doctor-authored reference explanation & Official English train/dev/test; four pool-overlap test questions removed for reporting & MG train family documented & 434 & 63 & 121 \\
MedExQA & 4-option MCQ & First member of headerless TSV explanation pair (column 6); human-validated reference & Official dev and test merged, deduplicated, then reshuffled & Not listed among documented MG post-training sources; other contamination is not excluded & 600 & 164 & 200 \\
PubMedQA & Yes/no/maybe & \texttt{long\_answer}; article abstract conclusion, not CoT & All PQA-L records reshuffled 600/200/200 & MG PubMedQA train family documented & 600 & 200 & 200 \\
MedMCQA & 4-option MCQ & \texttt{exp}; dataset expert-explanation field & Filtered official train records reshuffled into all three study splits & MG train family documented & 512 & 256 & 4,096 \\
\bottomrule
\end{tabular}
\end{table*}

AfriMed-QA contributes five-option, single-answer English items in our filtered view \citep{nimo2025afrimedqa}. We use the English MedExpQA configuration \citep{alonso2024medexpqa}; all retained rows contain its answer-reference-masked field. MedExQA's source TSV is headerless: columns 6 and 7 are the two explanation members, and the protocol always uses column 6 while retaining but never training on column 7 \citep{kim2024medexqa}. PubMedQA prepends abstract context to the research question and uses the article's conclusive abstract portion as auxiliary text \citep{jin2019pubmedqa}. MedMCQA uses its explanation field after removing records where that field is empty \citep{pal2022medmcqa}. Appendix~\ref{app:data-details} reports lengths, retained fractions, and exact split construction.

The normalized UID sets are disjoint, but MedExpQA contains repeated normalized question text: two keys cross pool--development and four cross pool--test; none cross development--test. For the main evaluation we remove every MedExpQA test row whose normalized question occurs in the pool, reducing its test set from 125 to 121. Prediction alignment is verified against length, answer-space width, and the complete ordered gold vector before positional filtering. All 36 evaluated method--seed selections include at least one pool item whose normalized question occurs in test. This post-hoc evaluation filter prevents duplicated test questions from contributing to metrics, but does not undo training on overlapping pool items or create a fully rebuilt deduplicated dataset. Removing the two pool-overlap development rows (63 to 61) leaves the selected coefficient unchanged at \(\lambda=448\) (Appendix~\ref{app:overlap-audit}). No exact normalized-question cross-split intersection was recovered for the other four study datasets.

The \dataset{MedGemma} technical report states that its post-training teacher responses and logits used training splits from MedMCQA, PubMedQA, MedExpQA, and AfriMed-QA \citep{sellergren2025medgemma}. Thus, four of the five benchmark families overlap the documented post-training sources. In particular, all study partitions for PubMedQA and MedMCQA were derived from source training splits, while AfriMed-QA retains its official test and MedExpQA retains official development and test splits. Without the unreleased \dataset{MedGemma} item manifest, we cannot compute exact example intersections or rule out broader public-data contamination. These experiments evaluate continued adaptation and subset selection relative to the model's medically post-trained state, not learning from wholly unseen medical domains; the model card itself recommends application-specific validation \citep{google2025medgemmacard}.

\subsection{Compared Selectors}

We group methods by whether candidate scoring reads auxiliary text; Appendix~\ref{app:baseline-table} gives an information-access table and implementation details. The auxiliary-text-blind group contains Entropy, Entropy-prefiltered Random (\randomfour{}), Margin, BALD-inspired hidden-state dropout, BatchBALD-inspired, Noise-Stability, SIFT-inspired, NAIT-inspired, and \method{}. \randomfour{} draws uniformly from the same top-\(\min(4B,N)\) entropy pool that \method{} reranks. Entropy instead deterministically keeps the most entropic items. BALD-inspired and BatchBALD-inspired use final-block-input hidden dropout rather than architectural or parameter dropout. Noise-Stability constructs mean-pooled sensitivity vectors followed by cosine k-center selection, and SIFT-inspired combines entropy with representation diversity. NAIT-inspired uses first-to-last prompt-activation deltas across decoder layers, oriented with development questions.

The auxiliary-text-aware group contains RSR-inspired, ASLEC-DROP-inspired, and TEMP-inspired. RSR-inspired uses a clipped rank--surprisal ratio; ASLEC-DROP-inspired removes the first token of heuristic reasoning steps from its likelihood score. TEMP-inspired fits an auxiliary direction from up to 64 development auxiliary-text--answer examples (63 for MedExpQA), then scores pool auxiliary texts under calibrated random and directional perturbations. These protocol-specific adaptations use the same downstream retained-set size but should not be interpreted as exact reproductions of the cited methods.

For BatchBALD-inspired, Noise-Stability, and SIFT-inspired, the nominal preselection is \(\min(5000,N)\); because every pool has \(N\le1500\), it equals the full pool here. Candidate-level auxiliary-text-blind methods do not read pool labels or auxiliary text. NAIT-inspired uses development questions without their labels, whereas \method{} uses labeled development outcomes to choose \(\lambda\). TEMP-inspired uses both labels and auxiliary text in its development direction.

\subsection{Training and Evaluation}

Each branch starts from \texttt{google/medgemma-4b-it} with no answer-only warm-up. The code did not pin a Hugging Face revision, so retained artifacts establish the repository ID but not the exact resolved model commit. We initialize a fresh rank-16 LoRA adapter with scale 32 and dropout 0.05 over \texttt{q/k/v/o/gate/up/down\_proj} modules \citep{hu2022lora}. Training uses two epochs, learning rate \(10^{-4}\), cosine decay, 3\% warm-up, fused AdamW, per-device batch size 1, gradient accumulation 8, and maximum length 1,024. No quantization is requested; loading uses bfloat16 when supported and otherwise float16, but the resolved precision and software package versions were not retained.

The exact training serialization is
\begin{quote}\small
\texttt{[question, options] \textbackslash nReasoning:\textbackslash n [auxiliary text]\\
Using the reasoning above, return only the final option letter.\\
Final Answer: [LETTER]}
\end{quote}
followed by EOS. Only the leading-space answer letter, newline, and EOS target tokens receive loss. Evaluation provides the question and options followed by the answer instruction, then chooses among legal next-letter logits; there is no free generation and therefore no invalid-output category. Accuracy and Macro-F1 use answer-option positions (A--C, A--D, or A--E) as classes, not medical diagnosis classes. We report mean and sample standard deviation across branch seeds 13, 23, and 37. For stochastic selectors, including \method{} and \randomfour{}, a branch seed determines both acquisition randomness and downstream training; fixed selections vary only in training. All adapted selectors share each dataset's fixed pool, development, and test partitions. Slurm manifests request GPUs labeled RTX5000 for \method{}, BALD-inspired, BatchBALD-inspired, Noise-Stability, SIFT-inspired, and \randomfour{}, and RTX6000 for NAIT-inspired, RSR-inspired, ASLEC-DROP-inspired, and TEMP-inspired. They record requested GRES classes rather than per-job GPU UUIDs, so the experiment families are not hardware-uniform or bitwise reproducible.

\paragraph{Archived no-update reference.}
We report a recovered single evaluation of \texttt{google/medgemma-4b-it} with no adapter or training. It uses the same documented question/options-only, restricted-letter evaluation protocol and the same four MedExpQA test exclusions. Its ordered UIDs match the audited eligible source records, and its gold-label order matches all three retained \method{} branches for each dataset; source and evaluator hashes were checked during the artifact audit. However, the older adapted outputs lack UID arrays, so gold-order agreement does not independently establish historical question/options identity. Neither code path pinned the model revision, and the surviving cache lacks the historical weights. Exact historical checkpoint equivalence therefore cannot be proven. We report this run as an archival reference, with no training-seed SD and no new inference, rather than as a fully matched before/after control.

\section{Results}
\label{sec:results}

Tables~\ref{tab:accuracy} and~\ref{tab:f1} preserve the complete evaluated comparison as a \emph{mixed-overlap benchmark audit}. For adapted selectors, the final column is the unweighted dataset mean within each seed, summarized over three seeds; the raw reference gives the unweighted mean of its single-run dataset values. The two tables report point estimates rather than a tournament with an equal tuning budget. MedExpQA metrics use the 121-item pool-overlap-filtered test set; other dataset values are unchanged.

\begin{table*}[t]
\centering
\caption{Test accuracy (\%). Adapted selectors use \(B=128\) and report mean \(\pm\) sample SD over three seed branches; Raw/no-update is one archived evaluation without a seed SD (Section~\ref{sec:experiments}). \randomfour{} denotes Entropy-prefiltered Random. \method{} uses dataset-specific development tuning; fixed comparators do not receive its 26-configuration search. Bold marks the largest selector point estimate, not statistical significance.}
\label{tab:accuracy}
\scriptsize
\setlength{\tabcolsep}{3.0pt}
\begin{tabular}{lrrrrrr}
\toprule
Method & AfriMed & MedExp & MedEx & PubMed & MedMCQA & \shortstack{Mean across\\datasets} \\
\midrule
\multicolumn{7}{l}{\emph{Archived reference without fine-tuning (one run)}} \\
Raw/no-update
& $43.27$ & $40.50$ & $36.00$ & $57.50$ & $34.86$ & $42.43$ \\
\midrule
\multicolumn{7}{l}{\emph{Auxiliary-text-blind candidate scoring}} \\
Entropy
& $48.69\!\pm\!0.34$ & $52.89\!\pm\!2.48$ & $50.33\!\pm\!1.76$ & $68.67\!\pm\!0.58$ & $35.94\!\pm\!1.91$ & $51.30\!\pm\!1.05$ \\
\randomfour{}
& $52.84\!\pm\!0.28$ & $50.69\!\pm\!3.91$ & $49.83\!\pm\!4.25$ & $71.00\!\pm\!1.00$ & $35.37\!\pm\!2.35$ & $51.95\!\pm\!0.64$ \\
Margin
& $51.79\!\pm\!1.87$ & $53.72\!\pm\!0.83$ & $50.17\!\pm\!4.80$ & $70.33\!\pm\!0.29$ & $37.77\!\pm\!1.33$ & $52.76\!\pm\!1.06$ \\
BALD-inspired
& $52.02\!\pm\!0.40$ & $52.34\!\pm\!1.26$ & $50.00\!\pm\!1.73$ & $69.67\!\pm\!0.29$ & $38.23\!\pm\!0.58$ & $52.45\!\pm\!0.39$ \\
BatchBALD-inspired
& $52.62\!\pm\!0.45$ & $50.69\!\pm\!5.98$ & $48.17\!\pm\!0.29$ & $70.33\!\pm\!0.58$ & $36.43\!\pm\!1.14$ & $51.65\!\pm\!1.14$ \\
Noise-Stability
& $49.55\!\pm\!1.36$ & $52.07\!\pm\!0.83$ & $46.33\!\pm\!5.01$ & $69.00\!\pm\!0.87$ & $39.30\!\pm\!2.71$ & $51.25\!\pm\!1.48$ \\
SIFT-inspired
& $52.39\!\pm\!1.97$ & $51.52\!\pm\!2.52$ & $45.33\!\pm\!1.04$ & $69.17\!\pm\!1.26$ & $37.51\!\pm\!4.48$ & $51.18\!\pm\!0.74$ \\
NAIT-inspired
& $49.55\!\pm\!2.16$ & $53.72\!\pm\!2.48$ & $52.00\!\pm\!3.04$ & $54.00\!\pm\!0.87$ & $36.08\!\pm\!1.00$ & $49.07\!\pm\!1.49$ \\
\method{} (dataset-tuned)
& $\mathbf{53.03\!\pm\!1.96}$ & $\mathbf{53.99\!\pm\!0.48}$ & $50.83\!\pm\!1.04$ & $70.00\!\pm\!1.32$ & $\mathbf{39.33\!\pm\!1.23}$ & $\mathbf{53.44\!\pm\!1.04}$ \\
\midrule
\multicolumn{7}{l}{\emph{Auxiliary-text-aware candidate scoring}} \\
RSR-inspired
& $50.64\!\pm\!3.59$ & $50.69\!\pm\!1.26$ & $\mathbf{52.50\!\pm\!3.50}$ & $69.00\!\pm\!1.00$ & $37.04\!\pm\!0.33$ & $51.97\!\pm\!0.41$ \\
ASLEC-DROP-inspired
& $52.47\!\pm\!1.40$ & $49.59\!\pm\!1.65$ & $49.17\!\pm\!3.06$ & $\mathbf{71.67\!\pm\!1.26}$ & $37.10\!\pm\!0.87$ & $52.00\!\pm\!0.73$ \\
TEMP-inspired
& $51.64\!\pm\!1.21$ & $51.79\!\pm\!2.90$ & $51.50\!\pm\!1.32$ & $70.33\!\pm\!0.76$ & $37.26\!\pm\!0.73$ & $52.51\!\pm\!0.40$ \\
\bottomrule
\end{tabular}
\par\smallskip
\parbox{0.98\textwidth}{\scriptsize \randomfour{} samples uniformly after entropy preselection. It equals full-pool uniform random for MedExpQA (\(N=434\)) and MedMCQA (\(N=512\)); full-pool random was not evaluated for AfriMed-QA, MedExQA, or PubMedQA.}
\end{table*}

\begin{table*}[t]
\centering
\caption{Test Macro-F1 (\%). Classes are answer-option positions; the single-run raw reference and adapted-selector reporting conventions match Table~\ref{tab:accuracy}.}
\label{tab:f1}
\scriptsize
\setlength{\tabcolsep}{3.0pt}
\begin{tabular}{lrrrrrr}
\toprule
Method & AfriMed & MedExp & MedEx & PubMed & MedMCQA & \shortstack{Mean across\\datasets} \\
\midrule
\multicolumn{7}{l}{\emph{Archived reference without fine-tuning (one run)}} \\
Raw/no-update
& $38.13$ & $37.12$ & $35.08$ & $31.75$ & $29.14$ & $34.24$ \\
\midrule
\multicolumn{7}{l}{\emph{Auxiliary-text-blind candidate scoring}} \\
Entropy
& $44.99\!\pm\!1.08$ & $50.31\!\pm\!3.59$ & $50.29\!\pm\!1.69$ & $48.56\!\pm\!0.43$ & $30.20\!\pm\!4.24$ & $44.87\!\pm\!1.53$ \\
\randomfour{}
& $49.61\!\pm\!0.48$ & $48.42\!\pm\!5.65$ & $49.74\!\pm\!4.08$ & $49.97\!\pm\!0.86$ & $32.81\!\pm\!0.49$ & $46.11\!\pm\!0.78$ \\
Margin
& $49.77\!\pm\!2.14$ & $51.24\!\pm\!1.23$ & $49.87\!\pm\!4.46$ & $49.34\!\pm\!0.23$ & $33.87\!\pm\!2.54$ & $46.82\!\pm\!1.00$ \\
BALD-inspired
& $48.17\!\pm\!0.56$ & $48.04\!\pm\!2.13$ & $49.82\!\pm\!1.54$ & $50.46\!\pm\!2.35$ & $35.65\!\pm\!1.72$ & $46.43\!\pm\!0.32$ \\
BatchBALD-inspired
& $49.48\!\pm\!0.21$ & $48.57\!\pm\!6.97$ & $48.20\!\pm\!0.23$ & $50.08\!\pm\!1.13$ & $32.74\!\pm\!1.52$ & $45.82\!\pm\!1.13$ \\
Noise-Stability
& $44.86\!\pm\!2.11$ & $48.80\!\pm\!0.49$ & $46.09\!\pm\!5.27$ & $50.50\!\pm\!3.34$ & $37.13\!\pm\!3.03$ & $45.48\!\pm\!1.54$ \\
SIFT-inspired
& $49.83\!\pm\!1.99$ & $48.12\!\pm\!3.01$ & $45.24\!\pm\!0.85$ & $49.39\!\pm\!1.14$ & $32.47\!\pm\!6.73$ & $45.01\!\pm\!0.75$ \\
NAIT-inspired
& $44.37\!\pm\!3.03$ & $49.70\!\pm\!2.84$ & $51.83\!\pm\!2.81$ & $24.29\!\pm\!1.83$ & $31.73\!\pm\!2.25$ & $40.39\!\pm\!1.74$ \\
\method{} (dataset-tuned)
& $\mathbf{50.79\!\pm\!2.45}$ & $\mathbf{51.72\!\pm\!0.95}$ & $50.55\!\pm\!0.76$ & $50.72\!\pm\!2.94$ & $\mathbf{37.29\!\pm\!1.10}$ & $\mathbf{48.21\!\pm\!0.74}$ \\
\midrule
\multicolumn{7}{l}{\emph{Auxiliary-text-aware candidate scoring}} \\
RSR-inspired
& $47.08\!\pm\!3.46$ & $46.66\!\pm\!1.88$ & $\mathbf{52.19\!\pm\!3.26}$ & $48.73\!\pm\!0.37$ & $33.51\!\pm\!2.22$ & $45.63\!\pm\!0.93$ \\
ASLEC-DROP-inspired
& $49.97\!\pm\!1.51$ & $47.16\!\pm\!1.88$ & $49.35\!\pm\!2.77$ & $\mathbf{53.16\!\pm\!2.30}$ & $34.16\!\pm\!3.05$ & $46.76\!\pm\!0.91$ \\
TEMP-inspired
& $47.61\!\pm\!1.80$ & $48.64\!\pm\!3.06$ & $51.33\!\pm\!1.49$ & $51.08\!\pm\!2.22$ & $31.74\!\pm\!1.89$ & $46.08\!\pm\!0.60$ \\
\bottomrule
\end{tabular}
\par\smallskip
\parbox{0.98\textwidth}{\scriptsize \randomfour{} is the pool-matched random comparator defined in Table~\ref{tab:accuracy}; it is not full-pool random on the three datasets with \(N>512\).}
\end{table*}

\subsection{RQ1: Overall Descriptive Performance}

In the mixed-overlap five-benchmark audit, dataset-specific development-tuned \method{} attains the highest observed mean accuracy (53.44\%) and Macro-F1 (48.21\%) among the evaluated selectors. The observed differences from Margin are +0.68 percentage points (pp) in accuracy and +1.40 pp in Macro-F1. \method{} has the largest point estimate on AfriMed-QA, filtered MedExpQA, and MedMCQA for both metrics. These are descriptive comparisons under unequal search flexibility and retrospective test reuse, not evidence that \method{} is statistically or universally superior.

The archived no-update reference averages 42.43\% accuracy and 34.24\% Macro-F1. The corresponding \method{} point estimates are higher by +11.01 pp and +13.97 pp. These descriptive archive differences provide absolute performance context, but the checkpoint and historical input-provenance limits in Section~\ref{sec:experiments} prevent a fully matched attribution to fine-tuning; no paired interval is reported for this comparison.

\subsection{RQ2: Pool-Matched Random Comparison}

\randomfour{} controls for the entropy prefilter and candidate-pool size, but does not equalize development-search flexibility. \method{} exceeds it descriptively by +1.49 pp in mean accuracy and +2.10 pp in mean Macro-F1. The direction is positive on 4/5 datasets for accuracy and 5/5 for Macro-F1. Table~\ref{tab:deltas} adds a paired bootstrap computed from retained predictions. Within each of 10,000 replicates, test items are resampled once per dataset and the three paired branch-seed identifiers are resampled; both draws are shared across methods. Intervals condition on these five fixed datasets and do not account for development tuning, benchmark choice, retrospective analysis decisions, or multiple exploratory comparisons.

\begin{table}[t]
\centering
\caption{Mean paired differences with 95\% percentile bootstrap intervals (pp). Direction counts are accuracy/Macro-F1. The intervals are descriptive, not a preregistered superiority test.}
\label{tab:deltas}
\small
\setlength{\tabcolsep}{4pt}
\begin{tabular}{lrrc}
\toprule
Comparison & Accuracy \(\Delta\) & Macro-F1 \(\Delta\) & Positive directions \\
\midrule
\method{} $-$ \randomfour{} & $+1.49\;[-0.48,3.54]$ & $+2.10\;[-0.04,4.37]$ & 4/5, 5/5 \\
\method{} $-$ Margin & $+0.68\;[-1.30,2.68]$ & $+1.40\;[-0.78,3.66]$ & 4/5, 5/5 \\
\bottomrule
\end{tabular}
\end{table}

For MedExpQA and MedMCQA, \(N\le4B\), so \randomfour{} is also full-pool uniform random. For AfriMed-QA, MedExQA, and PubMedQA, full-pool random was not evaluated; the experiment therefore does not isolate the value of entropy prefiltering on those datasets.

\subsection{RQ3: Dataset-Level Heterogeneity}

No method leads every column. On MedExQA, RSR-inspired has the largest observed accuracy (52.50\%) and Macro-F1 (52.19\%). On PubMedQA, ASLEC-DROP-inspired leads both metrics (71.67\% and 53.16\%), while \method{} accuracy (70.00\%) is also below \randomfour{} and several fixed selectors. PubMedQA selects \(\lambda=0\), so Drop contributes nothing to its final tuned score. Across datasets, selected \(\lambda\) ranges from 0 to 448; this scale coefficient has no demonstrated direct transferability.

Auxiliary-text-aware selectors are likewise heterogeneous: they lead MedExQA and PubMedQA but not all datasets or the five-dataset means. PubMedQA's sizable accuracy--Macro-F1 gap is consistent with class imbalance or class-specific errors, but the gap alone does not identify the cause. Overall, the results support \method{} as a tuning-dependent signal whose utility varies with the dataset.

As an overlap-aware sensitivity analysis, we exclude PubMedQA and MedMCQA, whose study test partitions were reconstructed from documented MedGemma post-training source training collections. Across AfriMed-QA, filtered MedExpQA, and MedExQA, \method{} averages 52.62\% accuracy and 51.02\% Macro-F1. Its differences from \randomfour{} are +1.50 pp and +1.76 pp (positive on 3/3 datasets for each metric), with paired intervals of \([-1.47,4.60]\) and \([-1.47,5.12]\). Differences from Margin are +0.73 pp and +0.73 pp, with intervals of \([-2.23,3.74]\) and \([-2.39,3.85]\). This subset is not claimed to be contamination-free; it only avoids test partitions directly reconstructed from the named source-training collections. Full method summaries appear in Appendix~\ref{app:overlap-sensitivity}.

\subsection{RQ4: Development-Search Cost and Scope}

The \(\lambda\) search contains 26 non-redundant dataset--coefficient configurations. Three seeds yield 78 development-evaluated adapter fits, versus 15 fits for a fixed selector evaluated once per dataset and seed. This is not an equal-search-budget comparison. Consequently, the observed differences cannot separate the value of the sensitivity score from the additional development search; a matched-search-budget comparison would be needed for that attribution. Counting only training-item presentations, the tuned search exposes
\[
78\times128\times2=19{,}968
\]
example-epochs. Training all 3,646 pool examples for two epochs and three seeds would expose \(3{,}646\times3\times2=21{,}876\), so the former is 91.27\% of the latter. This limited calculation omits scoring, evaluation, optimizer, sequence length, hardware, and scheduling overhead; it is neither a FLOP nor wall-clock comparison and does not establish end-to-end savings.

\section{Limitations and Scope}
\label{sec:limitations}

\paragraph{Causal scope.}
Every selected example includes auxiliary text during training, while evaluation omits it. We have no matched answer-only/no-auxiliary-text control, so the study does not show that auxiliary-text conditioning or CoT itself causes improvement. The archived no-update reference supplies absolute performance context, but exact historical checkpoint and input equivalence remain unresolved. Its descriptive gaps therefore do not isolate the effect of fine-tuning on a proven-identical starting model and test inputs. Training with auxiliary text and testing without it introduces a context shift. \randomfour{} is a pool-matched random comparator, but full-pool random is absent for AfriMed-QA, MedExQA, and PubMedQA; consequently, entropy prefiltering is not isolated there.

\paragraph{Selection validity.}
Dataset-specific \(\lambda\) tuning gives \method{} more search flexibility than fixed selectors and may overfit small development sets. Within the expanded run, development metrics selected \(\lambda\) before the frozen-winner test evaluation; nevertheless, the same test cohorts had already been evaluated in earlier analyses before that expanded grid was designed. The results are therefore retrospective exploratory estimates, not confirmation on never-viewed test data. Equation~\eqref{eq:rsp} is not guaranteed to be monotone in perturbation severity, and its fixed absolute \(\sigma\) is not normalized to hidden-state magnitude. We study one backbone, one \(B=128\) retained-set budget, one LoRA recipe, and three seeds. The paired intervals condition on the selected datasets and retained runs and do not repair these design limitations.

\paragraph{Data and generalization.}
Auxiliary provenance is heterogeneous: dataset-provided human explanations, doctor-authored reference text, an article conclusion, and an expert-explanation field are not interchangeable. The narrow preprocessing rules remove some explicit final-answer tails but do not constitute a comprehensive answer-leakage or factuality audit. MedExpQA has normalized question-text overlap between pool and development/test. We remove pool-overlap questions from its reported test metrics and audit the development winner, but this post-hoc filter is not a full pool/dev/test reconstruction or retraining; one within-test duplicate also remains. \dataset{MedGemma} documents post-training on four benchmark training families used here, and exact item overlap is unresolved. The five-dataset analysis therefore mixes backbone-overlap conditions. We have not demonstrated transfer across model revisions, other backbones, budgets, languages, institutions, or newly collected data. A confirmatory follow-up should deduplicate pool/development/test partitions before selection, freeze the protocol before inspecting a new test cohort, and evaluate both a similarly sized general-purpose backbone and medical datasets released after the evaluated models' documented training cutoffs. Such comparisons are not part of the present study, and dataset release dates alone would not prove absence of overlap.

\paragraph{Clinical scope.}
Medical-knowledge QA is not patient-level clinical evaluation. Accuracy and answer-position Macro-F1 do not evaluate calibration, harmful errors, explanation faithfulness, fairness, clinical utility, or deployment readiness. The model revision and software versions were not pinned in retained artifacts, and some selector branches used different GPU models, limiting bitwise reproducibility.

\paragraph{Ethics and safety.}
The experiments use previously released medical QA benchmarks and do not constitute patient-level clinical evaluation. \method{} does not assess whether auxiliary explanations are factual, faithful, unbiased, or safe; representation sensitivity may also prioritize ambiguous or mislabeled examples. Any use in clinical post-training would require data-governance review, expert validation, subgroup and calibration audits, and prospective evaluation. The resulting models should not be used directly for diagnosis, treatment, or patient management.

\section{Conclusion}

We studied fixed-budget selection of auxiliary-text-bearing examples for medical-QA post-training. In a retrospective mixed-overlap benchmark audit, dataset-specific development-tuned \method{} attains the highest observed five-dataset mean accuracy and Macro-F1 among the evaluated selectors, with modest descriptive differences over \randomfour{}. Results vary by dataset and do not establish statistical superiority, a causal benefit from auxiliary explanations, or clinical readiness. The archived no-update comparison gives absolute context, subject to unresolved historical checkpoint and input provenance. \method{} is therefore best viewed as an operational, tuning-dependent selection signal. Future work should investigate normalized and transferable sensitivity scores, additional backbones and budgets, explanation quality, and clinically relevant reliability evaluation.

\paragraph{Code and artifacts.}
The research-code repository is \url{https://github.com/woqingdoua/rsp-medical-qa-selection}. The accompanying retained-artifact bundle is described in Appendix~\ref{app:paired}.

\bibliographystyle{plainnat}
\bibliography{sample}

\clearpage
\appendix

\section{Reproducibility Details}
\label{app:reproducibility}

\subsection{Prompt, Loss Mask, and Truncation}

The protocol uses plain text rather than a chat template and tokenizes with \texttt{add\_special\_tokens=False}. The auxiliary-text-blind scoring/evaluation prompt is
\begin{quote}\small
\texttt{Please answer the following multiple-choice medical question:\\{}
[QUESTION]\\
A. [OPTION] ...\\
Choose one option. Return only the final option letter.\\
Final Answer:}
\end{quote}
Training inserts \texttt{Reasoning: [AUXILIARY TEXT]} between the options and answer instruction. Before tokenization, three regular expressions remove newline-delimited tails of the forms \texttt{final answer}, \texttt{therefore ... answer is}, or \texttt{correct option}; other answer cues are not masked. The target \texttt{" [LETTER]\textbackslash n"} and EOS are fully reserved and supervised. If necessary, the question/options prefix is truncated from its middle (preserving head and tail), then auxiliary text is right-truncated; the instruction and full target remain. Evaluation head--tail truncates a prompt to 1,024 tokens. Across the 15 final \method{} branches, all \(3\times128\) selected examples per dataset were retained and none required question or auxiliary-text truncation; this statement does not cover every comparator branch.

\subsection{Optimization and Model Loading}

LoRA uses rank 16, scale 32, dropout 0.05, no bias, and targets \texttt{q\_proj}, \texttt{k\_proj}, \texttt{v\_proj}, \texttt{o\_proj}, \texttt{gate\_proj}, \texttt{up\_proj}, and \texttt{down\_proj}. Training uses batch size 1, gradient accumulation 8, two epochs, fused AdamW, learning rate \(10^{-4}\), cosine scheduling, 3\% warm-up, gradient checkpointing, and seeds 13/23/37. The code requests \texttt{google/medgemma-4b-it} without an explicit revision and dynamically selects bfloat16 or float16 according to GPU support; no quantization is configured. Retained records do not establish exact Transformers/PEFT versions or resolved precision.

\subsection{Data Construction and Descriptive Statistics}
\label{app:data-details}

AfriMed-QA, MedExQA, and PubMedQA deduplicate lowercased, whitespace-normalized questions before their internal splits. AfriMed-QA filters official training/test data to single-answer five-option MCQs, shuffles the training portion with seed 20260826, and retains the official test. MedExpQA directly maps English train/validation/test to pool/dev/test without question deduplication. Across its 622 rows, 15 normalized keys cover 32 rows; the pool contains nine duplicated keys over 19 rows, the test contains one key over two rows, and exact key intersections are two for pool--development, four for pool--test, and zero for development--test. MedExQA merges its 25-item development collection with the 940-item test collection, removes one duplicate, then reshuffles to 600/164/200; its study test is therefore not an official untouched test. PubMedQA reshuffles the 1,000 labeled PQA-L records to 600/200/200 rather than preserving the publication's 500/500 protocol. MedMCQA filters official training records with nonempty \texttt{exp}, excludes 512 earlier-pilot UIDs, then samples 512/256/4,096 without an explicit question-deduplication step; the retained study rows nevertheless have no exact normalized-question duplicate or cross-split intersection.

\begin{table}[H]
\centering
\caption{Auxiliary-text length in whitespace-delimited words for each pool, and final retained fraction. These are not tokenizer-token counts.}
\label{tab:data-lengths}
\small
\begin{tabular}{lrrr}
\toprule
Dataset & Q1/median/Q3 words & \(B/N\) & Missing auxiliary text \\
\midrule
AfriMed-QA & 25/48/86 & 8.53\% & 0 \\
MedExpQA & 36/59.5/102.75 & 29.49\% & 0 \\
MedExQA & 64/84/103 & 21.33\% & 0 \\
PubMedQA & 27/37/49 & 21.33\% & 0 \\
MedMCQA & 28/57/96 & 25.00\% & 0 \\
\bottomrule
\end{tabular}
\end{table}

\section{MedExpQA Overlap Audit}
\label{app:overlap-audit}

The normalized-question key is the lowercased result of converting the question to text, stripping its ends, and collapsing each whitespace run to one space. Table~\ref{tab:medexp-overlap} was recomputed from the retained pool/development/test JSONL files rather than copied from manuscript prose. Anonymous SHA-256 UID hashes and row-level memberships accompany the source package; no question text is redistributed.

\begin{table}[H]
\centering
\caption{MedExpQA normalized-question overlap. Row counts include every implicated row, so they can exceed the key count.}
\label{tab:medexp-overlap}
\small
\begin{tabular}{lrrr}
\toprule
Split pair & Keys & Rows in first split & Rows in second split \\
\midrule
Pool--development & 2 & 2 & 2 \\
Pool--test & 4 & 5 & 4 \\
Development--test & 0 & 0 & 0 \\
\bottomrule
\end{tabular}
\end{table}

The pool has nine duplicated keys covering 19 rows; the original test has one duplicated key covering two rows. The main analysis removes the four test rows whose keys occur in the pool, yielding \(n=121\), but leaves the within-test duplicate because the prespecified correction targets cross-split leakage. A full first-occurrence reconstruction across pool, development, and test would instead produce 424/61/120 rows and require new selection and training; it was not performed. Across all 36 method--seed branches, the selected 128-item set contains one to four pool items whose key occurs in test; the union over development/test overlap contains one to five. Exact hashed memberships and selection signatures are supplied.

For development sensitivity, the two pool-overlap development rows are removed and all retained candidate predictions are rescored under the original rule. The nonredundant MedExpQA candidates are \(\lambda\in\{0,0.25,448,512,576\}\). As Table~\ref{tab:medexp-dev-sensitivity} shows, the winner remains 448, so no new fit or inference is needed.

\begin{table}[H]
\centering
\caption{MedExpQA development sensitivity after filtering 63 to 61 rows (mean across seeds, \%).}
\label{tab:medexp-dev-sensitivity}
\small
\begin{tabular}{rrr}
\toprule
\(\lambda\) & Accuracy & Macro-F1 \\
\midrule
0 & 56.28 & 56.38 \\
0.25 & 60.11 & 59.40 \\
448 & \textbf{61.20} & \textbf{60.80} \\
512 & 59.02 & 57.56 \\
576 & 59.56 & 57.58 \\
\bottomrule
\end{tabular}
\end{table}

\section{Backbone-Overlap-Aware Sensitivity}
\label{app:overlap-sensitivity}

Table~\ref{tab:three-dataset} restricts the descriptive average to benchmarks whose study test partitions were not directly reconstructed from documented MedGemma post-training source training collections: AfriMed-QA, filtered MedExpQA, and MedExQA. This label does not mean clean or contamination-free.

\begin{table}[H]
\centering
\caption{Three-dataset sensitivity (\%, mean \(\pm\) sample SD across seeds; equal dataset weight).}
\label{tab:three-dataset}
\small
\setlength{\tabcolsep}{4pt}
\begin{tabular}{lrr}
\toprule
Method & Accuracy & Macro-F1 \\
\midrule
Entropy & $50.64\!\pm\!1.49$ & $48.53\!\pm\!2.04$ \\
\randomfour{} & $51.12\!\pm\!1.35$ & $49.26\!\pm\!1.56$ \\
Margin & $51.89\!\pm\!1.77$ & $50.29\!\pm\!1.43$ \\
BALD-inspired & $51.45\!\pm\!0.83$ & $48.68\!\pm\!0.33$ \\
BatchBALD-inspired & $50.49\!\pm\!2.03$ & $48.75\!\pm\!2.20$ \\
Noise-Stability & $49.32\!\pm\!1.79$ & $46.59\!\pm\!2.07$ \\
SIFT-inspired & $49.75\!\pm\!0.95$ & $47.73\!\pm\!1.27$ \\
NAIT-inspired & $51.76\!\pm\!2.19$ & $48.63\!\pm\!2.55$ \\
\method{} (dataset-tuned) & $\mathbf{52.62\!\pm\!1.10}$ & $\mathbf{51.02\!\pm\!1.28}$ \\
RSR-inspired & $51.27\!\pm\!0.74$ & $48.64\!\pm\!1.17$ \\
ASLEC-DROP-inspired & $50.41\!\pm\!0.84$ & $48.83\!\pm\!0.57$ \\
TEMP-inspired & $51.64\!\pm\!1.03$ & $49.19\!\pm\!0.99$ \\
\bottomrule
\end{tabular}
\end{table}

\section{Selector Information and Adaptations}
\label{app:baseline-table}

\begin{table}[H]
\centering
\caption{Information access and protocol adaptation. ``Dev use'' distinguishes candidate scoring from auxiliary development procedures.}
\label{tab:information-access}
\scriptsize
\setlength{\tabcolsep}{1.8pt}
\begin{tabular}{p{2.2cm}p{3.0cm}cp{2.0cm}p{3.5cm}}
\toprule
Method & Selection signal & Reads pool auxiliary text? & Dev use & Adaptation in this study \\
\midrule
Entropy & Predictive entropy & No & None & Standard deterministic ranking \\
\randomfour{} & Uniform draw after entropy top-\(4B\) & No & None & Pool-matched random comparator \\
Margin & Negative top-two probability gap & No & None & Standard deterministic ranking \\
BALD-inspired & MI under hidden-state dropout & No & None & Dropout pre-hook on final-block input; not parameter dropout \\
BatchBALD-inspired & Greedy joint MI from eight dropout passes & No & None & Hidden-dropout posterior proxy; 1,000 joint-label samples \\
Noise-Stability & Sensitivity vector plus cosine k-center & No & None & Study-defined operational comparator; relative RMS noise 0.03 \\
SIFT-inspired & Entropy plus cosine-distance diversity & No & None & Fixed diversity weight 0.5 \\
NAIT-inspired & Layerwise first-to-last activation delta & No & Unlabeled dev questions & One oriented PCA direction per layer; shared selection, three training seeds \\
\method{} & Entropy then Eq.~\eqref{eq:rsp} & No & Dev labels choose \(\lambda\) & Absolute Gaussian final-block-output perturbation \\
\midrule
RSR-inspired & Negative clipped mean-rank/surprisal ratio & Yes & None & Token ranks clipped at 100 \\
ASLEC-DROP-inspired & Rationale likelihood after step-first-token removal & Yes & None & Heuristic newline/sentence steps, at most 48 \\
TEMP-inspired & Loss under random/directional checkpoint perturbations & Yes & Dev labels and text & PEFT direction from up to 64 dev examples; calibrated perturbations \\
\bottomrule
\end{tabular}
\end{table}

BALD-inspired uses eight elementwise inverted-Bernoulli passes with probability 0.1 at the input to the final decoder block. BatchBALD-inspired reuses those distributions, preselects the top \(\min(5000,N)\) BALD scores, and greedily constructs a batch. Noise-Stability first randomly orders the same \(\min(5000,N)\) pool, adds eight per-example RMS-scaled perturbations at the final-block input, derives a normalized hidden-sensitivity vector, and applies seeded cosine k-center. SIFT-inspired uses mean-pooled final hidden representations with entropy and cosine diversity. Because \(N\le1500\), each 5,000-item cap is inactive.

NAIT-inspired computes, for every decoder block, the difference between the last prompt position and first valid token, estimates one oriented principal component from development questions, and sums candidate projections across layers. The acquisition calculation uses one fixed PCA seed; its reported variability comes from three downstream training seeds on that selection. RSR-inspired and ASLEC-DROP-inspired read pool auxiliary text but not pool answer labels. TEMP-inspired uses 63 labeled development explanation examples for MedExpQA and 64 for each other dataset, then reads pool explanatory text without pool labels. It uses a parameter-efficient direction and corrected bracket/bisection calibration rather than the original method's full-parameter trajectory.

All exported selector scores are consumed in descending order, with original row index breaking exact ties. For BALD-inspired, larger mutual information is preferred. BatchBALD-inspired, Noise-Stability, and SIFT-inspired encode earlier greedy choices with larger rank scores; Noise-Stability is therefore a diversity selector over sensitivity vectors, not a rule that directly prefers either lower or higher raw noise change. NAIT-inspired prefers a larger summed oriented-PC projection. RSR-inspired negates its clipped mean-rank/mean-surprisal ratio, so a smaller raw ratio is preferred; ranks are clipped at 100 and rationale context at 384 tokens. ASLEC-DROP-inspired prefers larger mean log probability after removing each heuristic step's first token, with at most 48 steps and 384 rationale tokens. TEMP-inspired gives priority to larger directional loss increase within its clustering procedure; the canonical branch uses four checkpoints, four examples per cluster, at most ten calibration evaluations, random loss-ratio target 2--3, and directional target 2.5--3.5. Its auxiliary rank-16 LoRA direction is trained for one epoch at learning rate \(10^{-4}\).

\section{Development-Search Accounting}
\label{app:cost}

\begin{table}[H]
\centering
\caption{Verified adapter-fit and evaluation accounting. The \method{} row counts all non-redundant development candidates; its test column counts only frozen winners.}
\label{tab:cost-audit}
\small
\begin{tabular}{lrrrr}
\toprule
Selector and scope & Branches & Fits & Dev evals & Test evals \\
\midrule
\randomfour{} (five datasets) & 15 & 15 & 15 & 15 \\
One fixed selector (five datasets) & 15 & 15 & 15 & 15 \\
\method{} \(\lambda\)-grid & 78 & 78 & 78 & 15 \\
\bottomrule
\end{tabular}
\end{table}

For \randomfour{}, nine branches (AfriMed-QA, MedExQA, and PubMedQA) come from the dedicated entropy-prefiltered-random run. Six earlier full-pool random branches are reused for MedExpQA and MedMCQA because \(N\le4B\), making the definitions identical. No \(\lambda\)-grid evaluations are attributed to this comparator. The \method{} search reuses 30 fits from the initial grid and adds 48 fits for the expanded grid, for 78 total development-evaluated fits. Only the 15 dataset--seed winners are tested.

\subsection{Expanded \texorpdfstring{\(\lambda\)}{lambda} Grid}
\label{app:lambda-grid}

The union of initial and expanded values is
\begin{equation}
\begin{split}
\lambda\in\{&0,0.25,16,24,32,40,48,56,64,80,100,\\
             &128,192,256,320,384,448,512,576,768,1024\}.
\end{split}
\end{equation}
Selections with identical ordered UID signatures are evaluated once. All settings for a dataset--seed pair reuse its saved Flip and Drop components.

For one pass, a flip contributes \(1/K\), whereas a non-flipping contraction \(\delta\) contributes \(\lambda\delta/K\). The latter can exceed the former if \(\lambda\delta>1\). Using saved clean margins, Table~\ref{tab:score-scale} reports how often the \emph{maximum possible} contraction \(m_i^{(0)}\) satisfies this inequality; it does not say that the observed noisy contraction reached that maximum. Identical counts were recovered from all three seed score files.

\begin{table}[H]
\centering
\caption{Candidate-level scale audit within the entropy-restricted pool.}
\label{tab:score-scale}
\small
\begin{tabular}{lrrr}
\toprule
Dataset & Selected \(\lambda\) & \(\lambda m_i^{(0)}>1\) & Fraction \\
\midrule
AfriMed-QA & 0.25 & 0/512 & 0.00\% \\
MedExpQA & 448 & 429/434 & 98.85\% \\
MedExQA & 24 & 499/512 & 97.46\% \\
PubMedQA & 0 & 0/512 & 0.00\% \\
MedMCQA & 56 & 505/512 & 98.63\% \\
\bottomrule
\end{tabular}
\end{table}

\section{Per-Seed Test Results}
\label{app:per-seed}

Tables~\ref{tab:per-seed-acc} and~\ref{tab:per-seed-f1} expose the values underlying the mean and sample SD in the main tables. Each cell is ordered as branch seeds 13/23/37; stochastic acquisitions and training share that seed as described in Section~\ref{sec:experiments}. The canonical sources are the expanded-development \method{} family for the fixed auxiliary-text-blind selectors, the latest-protocol family for RSR-inspired/ASLEC-DROP-inspired/NAIT-inspired, the corrected TEMP family, and the dedicated \randomfour{} audit.

\begin{table}[H]
\centering
\caption{Per-seed test accuracy (\%); order 13/23/37.}
\label{tab:per-seed-acc}
\scriptsize
\setlength{\tabcolsep}{2.2pt}
\begin{tabular}{lccccc}
\toprule
Method & AfriMed & MedExp & MedEx & PubMed & MedMCQA \\
\midrule
Entropy & 48.77/48.99/48.32 & 55.37/52.89/50.41 & 52.00/50.50/48.50 & 69.00/69.00/68.00 & 37.18/33.74/36.89 \\
\randomfour{} & 53.14/52.80/52.58 & 46.28/52.07/53.72 & 53.00/45.00/51.50 & 71.00/72.00/70.00 & 33.13/37.82/35.16 \\
Margin & 53.48/49.78/52.13 & 53.72/52.89/54.55 & 54.50/51.00/45.00 & 70.50/70.00/70.50 & 37.65/36.50/39.16 \\
BALD-inspired & 51.91/52.47/51.68 & 53.72/51.24/52.07 & 51.00/48.00/51.00 & 69.50/69.50/70.00 & 37.82/38.89/37.99 \\
BatchBALD-inspired & 53.03/52.69/52.13 & 53.72/54.55/43.80 & 48.00/48.00/48.50 & 71.00/70.00/70.00 & 35.13/36.89/37.28 \\
Noise-Stability & 50.34/50.34/47.98 & 52.07/51.24/52.89 & 46.00/51.50/41.50 & 68.00/69.50/69.50 & 36.72/42.11/39.06 \\
SIFT-inspired & 50.11/53.48/53.59 & 52.07/48.76/53.72 & 44.50/46.50/45.00 & 70.50/69.00/68.00 & 42.68/34.79/35.06 \\
NAIT-inspired & 51.12/47.09/50.45 & 56.20/53.72/51.24 & 55.50/50.00/50.50 & 55.00/53.50/53.50 & 36.13/37.06/35.06 \\
\method{} (dataset-tuned) & 52.91/55.04/51.12 & 53.72/54.55/53.72 & 50.00/52.00/50.50 & 68.50/71.00/70.50 & 39.18/40.62/38.18 \\
RSR-inspired & 46.52/53.14/52.24 & 50.41/52.07/49.59 & 56.50/51.00/50.00 & 70.00/68.00/69.00 & 36.99/37.40/36.74 \\
ASLEC-DROP-inspired & 51.35/54.04/52.02 & 49.59/47.93/51.24 & 52.50/46.50/48.50 & 73.00/70.50/71.50 & 37.48/37.72/36.11 \\
TEMP-inspired & 53.03/50.78/51.12 & 52.07/48.76/54.55 & 52.50/52.00/50.00 & 69.50/71.00/70.50 & 37.40/37.92/36.47 \\
\bottomrule
\end{tabular}
\end{table}

\begin{table}[H]
\centering
\caption{Per-seed test Macro-F1 (\%); order 13/23/37.}
\label{tab:per-seed-f1}
\scriptsize
\setlength{\tabcolsep}{2.2pt}
\begin{tabular}{lccccc}
\toprule
Method & AfriMed & MedExp & MedEx & PubMed & MedMCQA \\
\midrule
Entropy & 45.23/45.93/43.81 & 53.54/50.95/46.45 & 51.95/50.37/48.56 & 48.56/48.99/48.12 & 33.70/25.49/31.41 \\
\randomfour{} & 50.12/49.52/49.18 & 42.00/50.66/52.62 & 52.76/45.10/51.37 & 50.02/50.81/49.08 & 32.41/33.35/32.66 \\
Margin & 51.55/47.40/50.36 & 50.42/50.63/52.66 & 53.85/50.73/45.04 & 49.49/49.46/49.08 & 34.07/31.23/36.29 \\
BALD-inspired & 47.59/48.19/48.72 & 48.93/49.59/45.61 & 50.59/48.04/50.82 & 53.17/49.09/49.12 & 33.67/36.50/36.76 \\
BatchBALD-inspired & 49.58/49.25/49.62 & 51.45/53.64/40.63 & 48.05/48.08/48.46 & 51.39/49.48/49.38 & 31.65/32.09/34.47 \\
Noise-Stability & 46.89/45.00/42.69 & 48.41/48.65/49.36 & 45.97/51.42/40.89 & 47.08/50.67/53.75 & 34.72/40.53/36.13 \\
SIFT-inspired & 47.58/51.35/50.57 & 47.34/45.58/51.45 & 44.53/46.17/45.02 & 49.59/50.41/48.15 & 40.24/28.50/28.66 \\
NAIT-inspired & 46.29/40.88/45.94 & 52.96/48.45/47.70 & 55.07/50.01/50.42 & 26.41/23.24/23.24 & 31.27/34.18/29.75 \\
\method{} (dataset-tuned) & 51.40/52.87/48.09 & 51.16/52.81/51.18 & 50.03/51.42/50.20 & 48.29/49.88/53.99 & 37.45/38.30/36.13 \\
RSR-inspired & 43.40/50.26/47.59 & 46.05/48.77/45.17 & 55.93/50.65/49.97 & 49.15/48.59/48.46 & 35.97/32.93/31.65 \\
ASLEC-DROP-inspired & 49.09/51.72/49.11 & 46.33/45.85/49.31 & 52.40/47.00/48.66 & 53.69/55.15/50.64 & 34.91/36.76/30.81 \\
TEMP-inspired & 49.68/46.39/46.77 & 48.07/45.90/51.95 & 52.36/52.02/49.62 & 53.65/49.89/49.70 & 29.61/33.21/32.40 \\
\bottomrule
\end{tabular}
\end{table}

\section{Paired-Recomputation Artifacts}
\label{app:paired}

The paired analysis is deterministic and uses retained \texttt{test\_probs} and \texttt{test\_gold} arrays; it performs no training or inference. For each bootstrap replicate and dataset, one test-item resample is shared across methods and a three-draw resample of paired branch-seed identifiers is shared across methods. Metrics are averaged over the resampled seeds and then equally over the fixed datasets. Macro-F1 retains each dataset's full answer-position label space even if a resample omits a label. Original NPZ files do not retain test UIDs, so positional pairing relies on the common frozen split order; before filtering, the recomputation verifies test length, answer-space width, and the complete ordered gold vector for every method and seed against the hashed source test manifest. The anonymous bundle then stores the verified arrays, hashed UID/order audit, exact configuration, overlap reports, leakage-filtered metrics, five- and three-dataset bootstrap summaries, and the analysis script. It contains neither raw question text nor model weights.

\end{document}